\documentclass[a4paper]{article}

\usepackage{enumitem}
\usepackage[utf8]{inputenc}
\usepackage[T1]{fontenc}
\usepackage{textcomp}
\usepackage{lmodern}
\usepackage{babel}
\usepackage{lipsum}
\usepackage{graphicx}
\usepackage{scalerel,stackengine,amsmath}
\usepackage{subcaption}
\usepackage{tikz}
\usetikzlibrary{positioning, calc}
\usepackage{algorithm}
\usepackage{algpseudocode}
\usepackage{accents}
\usepackage{xcolor,color}
\usepackage{parskip}
\usepackage{tikz,varwidth}
\usetikzlibrary{calc,shadings,shapes,fit,arrows,positioning}

\usepackage{bm}
\usepackage{mathrsfs,empheq}
\usepackage{amssymb,amsmath,amsthm,amsfonts,mathtools}
\usepackage[hyperfootnotes=false]{hyperref}
\usepackage{thmtools}
\usepackage{stmaryrd}
\usepackage{multirow}
\usepackage{booktabs}
\usepackage{alphabeta} % Allows \alpha, etc. in text mode.
\usepackage{thm-restate}
\usepackage{nicefrac}

\DeclareMathOperator*{\argmax}{\mathrm{arg\,max}}

\DeclareMathOperator*{\agg}{\sf agg}

\newtheorem{theorem}{Theorem}[section]

\newtheorem{definition}[theorem]{Definition}

\newif\ifarxiv
\arxivfalse

\newif\ifopus
\opustrue

\ifarxiv
    \newif\ifarxivoropus
        \arxivoropustrue
\else
    \ifopus
        \newif\ifarxivoropus
            \arxivoropustrue
    \else
        \newif\ifarxivoropus
            \arxivoropusfalse
    \fi
\fi

\usepackage[scaled=0.95]{FiraMono}
\ifarxiv
\usepackage{arxiv}
\usepackage{charter}
\usepackage{fullpage}
\fi

\ifopus
\usepackage{zibtitlepage}
\usepackage{arxiv}
\usepackage{charter}
\usepackage{fullpage}
\fi

\hypersetup{
    colorlinks=true,
    linkcolor=blue,
    filecolor=magenta,
    urlcolor=cyan,
    citecolor=blue
    }

\ifarxivoropus
\author{
    \href{https://orcid.org/0000-0003-3673-966X}{\includegraphics[scale=0.06]{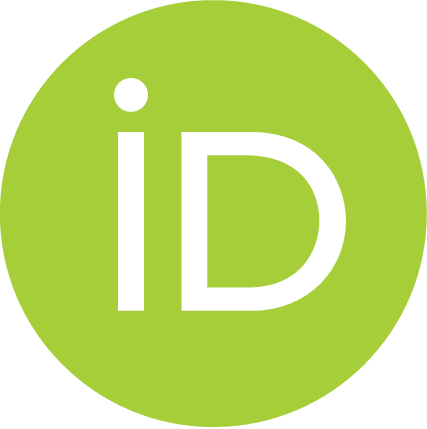}}\hspace{1mm}Gabriele Iommazzo\thanks{LIX CNRS, \'Ecole Polytechnique, Institut Polytechnique de Paris, Palaiseau, France. Gabriele Iommazzo is now at Zuse Institute Berlin, Berlin, Germany (\texttt{iommazzo@zib.de})}\\
    \And
    \href{https://orcid.org/0000-0002-4040-0960}{\includegraphics[scale=0.06]{orcid_id_icon.eps}}\hspace{1mm}Claudia D'Ambrosio\footnotemark[2]\\
    \texttt{dambrosio@lix.polytechnique.fr} \\
    \And
    \href{https://orcid.org/0000-0002-5704-3170}{\includegraphics[scale=0.06]{orcid_id_icon.eps}}\hspace{1mm}Antonio Frangioni\thanks{Dip.~di Informatica, Universit\`a di Pisa, Pisa, Italy} \\
    \texttt{frangio@di.unipi.it} \\
    \And
    \href{https://orcid.org/0000-0003-3139-6821}{\includegraphics[scale=0.06]{orcid_id_icon.eps}}\hspace{1mm}Leo Liberti\footnotemark[2]\\
    \texttt{liberti@lix.polytechnique.fr} \\
}
\else
\author{Gabriele Iommazzo\inst{1}\orcidID{0000-0003-3673-966X} \and
Claudia D'Ambrosio\inst{2}\orcidID{0000-0002-4040-0960}\and
Antonio Frangioni\inst{3}\orcidID{0000-0002-5704-3170}\and
Leo Liberti\inst{2}\orcidID{0000-0003-3139-6821}
}
\authorrunning{G. Iommazzo et al.}
\institute{LIX CNRS, \'Ecole Polytechnique, Institut Polytechnique de Paris, Palaiseau, France\and
Dip.~di Informatica, Università di Pisa, Pisa, Italy\\
\email{iommazzo@zib.de, \{dambrosio,giommazz,liberti\}@lix.polytechnique.fr}}
}
\fi

\ifopus
\ZTPAuthor{
	\ZTPHasOrcid{Gabriele Iommazzo}{0000-0003-3673-966X},
	\ZTPHasOrcid{Claudia D'Ambrosio}{0000-0002-4040-0960},
	\ZTPHasOrcid{Antonio Frangioni}{0000-0002-5704-3170},
	\ZTPHasOrcid{Leo Liberti}{0000-0003-3139-6821},}
\ZTPTitle{A learning-based mathematical programming formulation for the automatic configuration of optimization solvers}
\ZTPNumber{23-28}
\ZTPMonth{December}
\ZTPYear{2023}
\fi

\title{The Algorithm Configuration Problem}

\begin{document}

% \ifopus
% \zibtitlepage
% \fi

\maketitle

\vspace{5mm}

\begin{center}
\begin{minipage}{0.85\textwidth}
\begin{center}
 \textbf{Abstract}
\end{center}
{\small
The field of algorithmic optimization has significantly advanced with the development of methods for the automatic configuration of algorithmic parameters. This article delves into the Algorithm Configuration Problem, focused on optimizing parametrized algorithms for solving specific instances of decision/optimization problems. We present a comprehensive framework that not only formalizes the Algorithm Configuration Problem, but also outlines different approaches for its resolution, leveraging machine learning models and heuristic strategies. The article categorizes existing methodologies into per-instance and per-problem approaches, distinguishing between offline and online strategies for model construction and deployment. By synthesizing these approaches, we aim to provide a clear pathway for both understanding and addressing the complexities inherent in algorithm configuration.
% We would like to encourage you to list your keywords within
% the abstract section using the \keywords{...} command.
}
\end{minipage}
\end{center}

\section{Introduction}\label{s:acp_intro}

Automatic configuration of algorithmic parameters is an area of active research \cite{pac_asp,acp_survey_ker}, going back to the foundational work \cite{rice_acp}, published in 1976.
The Algorithm Configuration Problem (ACP) focuses on parametrized algorithms, deployed to solve instances of a given decision or optimization problem. It concerns the identification of algorithmic parameters yielding the best performance when the algorithm is run on its input.

Formally, we let $\mathcal{A}$ be a configurable target algorithm. Its input consists of an instance of the problem being solved and an array of parameters; we call the latter ``algorithmic configuration''.
\par
The inputs of the ACP are:
\begin{itemize}
\item $\Pi$: the decision/optimization problem to be solved by $\mathcal{A}$, consisting of a (potentially) infinite set of instances.
Each instance is typically {described by} a string in an appropriate alphabet, most often an ASCII or binary file encoding the (vectors of) discrete/continuous values that characteri{z}e the instance data. {We will assume that $\Pi$ is endowed with a metric and that there exists a function $\mathsf{dist}(\cdot, \cdot) \in \mathbb{R}$, which can measure the distance between points in $\Pi$ and approaches zero if two instances are similar. The function $\mathsf{dist}$ is not necessarily unique, but we will assume that it is always defined by an expression in closed algebraic form;}
\item $\mathcal{C}_{\mathcal{A}}$: the set of parameter configurations of $\mathcal{A}$, each usually encoded by vectors of $q$ continuous and/or discrete/categorical values representing parameters of different types (boolean, numeric, categorical). Not all possible parameter values may be admissible, due to constraints on a parameter domain or logical conditions concerning multiple parameters. Thus, for simplicity, we assume that $\mathcal{C}_{\mathcal{A}}$ only contains feasible algorithmic configurations;
\item $p_{\mathcal{A}}$: the performance function
\begin{equation}
	p_{\mathcal{A}}: \Pi \times \mathcal{C}_{\mathcal{A}}
	\longrightarrow \mathbb{R}
	\label{eq:perf_fun}
\end{equation}
of $\mathcal{A}$,mapping a pair $(\pi,c)$ (instance, parameter configuration) to the outcome of running $\mathcal{A}$, configured by $c$, to solve the instance $\pi$. The encoding of $p_\mathcal{A}$ is a single continuous or discrete value. The performance $p_\mathcal{A}$ is either a cost measure (e.g., CPU time, number of iterations performed, etc.) or a quality measure (e.g., the accuracy achieved by a Machine Learning (ML) predictor at a certain iteration of the training process, the integrality gap reported by an optimization solver within a certain time limit, etc.). Depending on the case at hand, one aims at appropriately minimizing or maximizing it.
\end{itemize}
With the specifications given above, the ACP is formally defined as follows:

\begin{definition}[ACP]
Given a tuple $(\mathcal{A}, \pi, p_\mathcal{A})$, $\pi \in \Pi$, find the algorithmic configuration $c^\ast_{\pi} \in \mathcal{C}_\mathcal{A}$ providing the optimal performance $p_\mathcal{A}$ for solving $\pi$ with $\mathcal{A}$.
\label{d:acp}
\end{definition}

A variant of the ACP is the Algorithm Selection Problem (ASP), where one seeks to pick, from a given set of configured algorithms, the best one for solving a specific instance. However, one can see the choice of which algorithm to pick as the one single parameter of a meta-algorithm for solving $\Pi$, and therefore the ASP is a special case of the ACP. The ACP is generally very hard both in theory and in practice, because algorithms may have a large number of configurable parameters and a closed-form algebraic expression of $p_\mathcal{A}$ is almost never available. Yet, the ACP has a large number of very relevant applications, such as the configuration of constraint programming or mathematical optimization solvers (see, e.g., \cite{baron_tuning}), the hyperparameter tuning of ML pipelines, the administration of ad-hoc medical treatments, and many others; the interested reader is referred, e.g.,to \cite{ac_best_practices,acp_survey_ker} and the references therein for a more detailed treatment of the subject.

In this article, we supply a comprehensive framework for describing any approach to the ACP: we identify the core building components for designing an ACP solution strategy, and discuss their possible use patterns and implementation.

In Tab.~\ref{shorthand_acp}, we recall the abbreviations used throughout the article:

\hypertarget{acp_anchor}{
	\begin{table}[ht]
		\centering
		\begin{tabular}{c|c}
			extended name & shorthand \\
			\hline
			Algorithm Configuration Problem & ACP \\
			Machine Learning & ML \\
			knowledge-encoding process & K-EP\\
		\end{tabular}
		\caption{Recurring abbreviations\label{shorthand_acp}}
	\end{table}
}

%---------------------------------------------
%---------------------------------------------
\section{Definitions}\label{s:acp_schema}

Research efforts in the ACP literature have focused on developing methodologies to solve it efficiently in an automated fashion. Although this has been implemented in several different ways, all approaches share the same goal, i.e., the construction of a {\it recommender}:

\begin{definition}[recommender]
 The recommender is a function
 \begin{equation}
  \Psi_\mathcal{M}: \Pi \to \mathcal{C}_\mathcal{A}
  \label{eq:acp_recommender}
 \end{equation}
which, given an instance $\pi \in \Pi$, is capable of selecting a configuration ${\bar{c}^\ast_\pi} \in \mathcal{C}_\mathcal{A}$ for solving $\pi$ more efficiently than with other configurations of $\mathcal{A}$.  \label{d:acp_fun}
\end{definition}
In other words, a recommender is a heuristic for the ACP, and ${\bar{c}^\ast_\pi}$ is, hopefully, a good approximation of the optimal configuration for solving $\pi$, with respect to the performance function $p_\mathcal{A}$.
The fundamental requirement of $\Psi_\mathcal{M}$ is that it should be able to produce its output in a ``short'' time, so as not to offset the advantages of choosing a better configuration.

Our notation underlines the fact that the structure of $\Psi_\mathcal{M}$ is usually determined by a model $\mathcal{M}$ encoding some knowledge about $p_\mathcal{A}$. In fact, an important phase of all ACP methodologies is devoted to the construction of $\mathcal{M}$. It is in general difficult (if at all possible) to build accurate enough analytical models of the performances of complex algorithms. Thus, most practical $\mathcal{M}$ are ``data-driven'', in the sense that they are constructed from experiments. For the recommender to be able to choose the right ${\bar{c}^\ast_\pi}$ for a given $\pi$, it should be able to assess $p_\mathcal{A}$ anywhere on the set $\Pi \times \mathcal{C}_\mathcal{A}$. However, an exhaustive evaluation of $p_\mathcal{A}$ over $\Pi \times \mathcal{C}_\mathcal{A}$ is almost always impossible in practice: $\Pi$ is an infinite set, and $\mathcal{C}_\mathcal{A}$ usually grows exponentially in the number of parameters, which can be large. Furthermore, since $p_\mathcal{A}$ itself is typically a black-box function (i.e., it has no analytic form), the only way to evaluate it is to directly run $\mathcal{A}$, which can be extremely costly. Therefore, a significant component of ACP approaches is how the set $\Pi \times \mathcal{C}_\mathcal{A}$ is explored to build $\mathcal{M}$.

Given the difficulty of assessing algorithmic performance{s} on $\Pi \times \mathcal{C}_\mathcal{A}$ exhaustively, the construction of $\Psi_\mathcal{M}$ in Eq.~\eqref{eq:acp_recommender} {
%\sout{always}
typically} involves the selection of sets $\Pi' \subset \Pi$ and $\mathcal{C}'_\mathcal{A} \subseteq \mathcal{C}_\mathcal{A}$.
Of these, $\Pi'$ is meant to be ``representative'' of $\Pi$, usually in the sense that it preserves the characteristics and information of $\Pi$. Sometimes, this can instead be taken as the fact that $\Pi'$ contains the most ``difficult'' instances for $\mathcal{A}$, in that all others are solved efficiently {and quickly}, and do not require a dedicated algorithmic configuration. Since there is no automatic way of choosing $\Pi'$, most ACP approaches take it as given and rely on existing libraries, hand-picked by problem experts. Therefore, we assume that $\Pi'$ is always available, or can be easily generated for deploying an ACP methodology. Further, we assume that $\Pi'$ is specified before the construction of $\mathcal{M}$, and never updated.
As for the selection of $\mathcal{C}'_\mathcal{A}$, we assume that the algorithmic performance of different configurations is typically unknown before launching an ACP approach, otherwise, there would be no need to even construct a recommender $\Psi_\mathcal{M}$.
%{\sout{This means that $\mathcal{C}'_\mathcal{A}$ is often selected during the construction of $\Psi_\mathcal{M}$, instead of being picked {\it a priori}, as in the case of $\Pi'$; the (partly constructed) model $\mathcal{M}$ can also be useful in this context.}}
{The set of parameters to tune, and the corresponding $\mathcal{C}'_\mathcal{A}$, are sometimes picked \textit{a priori}, as in the case of $\Pi'$. This choice is typically supported by domain experts, who have extensive experience working with $\mathcal{A}$ or in-depth knowledge of $\Pi$, and might be able to assess which subset of parameters --- or, sometimes, even parameter configurations --- are most likely to influence algorithmic performances.
Sometimes, instead, $\mathcal{C}'_\mathcal{A}$ is selected during the construction of $\Psi_\mathcal{M}$; the (partly constructed) model $\mathcal{M}$ can also be useful in this context.}

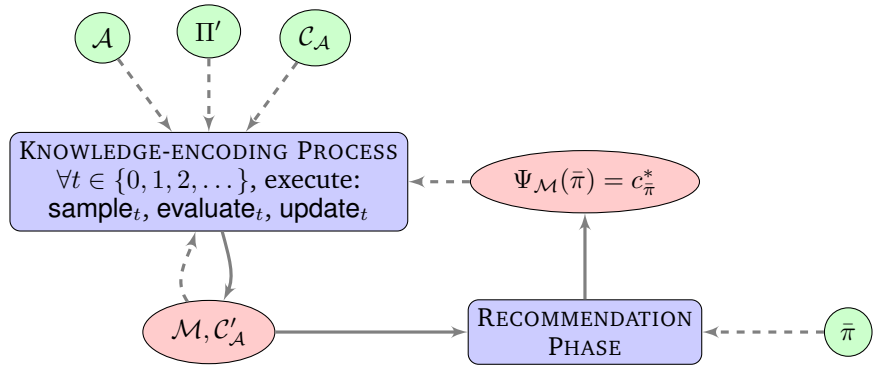
\begin{figure}[ht]
    \centering
    \begin{tikzpicture}[node distance=2cm, auto]
        \tikzstyle{block} = [draw, rectangle, fill=blue!20, text centered, rounded corners]
        \tikzstyle{cloud} = [draw, ellipse,fill=green!20, text centered]
        \tikzstyle{line} = [draw, very thick, color=black!50, -latex']
        \tikzstyle{dh-line} = [draw, very thick, color=black!50, latex'-latex'] % double-headed arrow
        % Place nodes
        \node [block, align=center] (lproc) {\sc {Knowledge-encoding Process}\\$\forall t \in \{0, 1, 2, \dots\}$, execute:\\
            {\sf sample$_t$}, {\sf evaluate$_t$}, {\sf update$_t$}};
        %%%%
        \node [cloud, above left of=lproc, yshift=5mm] (A) {$\mathcal{A}$};
        \node [cloud, above of=lproc] (Pi) {$\Pi'$};
        \node [cloud, above right of=lproc, yshift=5mm] (C) {$\mathcal{C}_\mathcal{A}$};
        \node [cloud, below of=lproc, fill=red!20] (M) {$\mathcal{M}, {\mathcal{C}'_\mathcal{A}}$};
        \node [block, right of=M, text centered, xshift=3cm, align=center] (recomm) {{\sc Recommendation}\\{\sc Phase}};
        \node [cloud, right of=recomm, xshift=1.5cm] (pi) {$\bar{\pi}$};
        \node [cloud, above of=recomm, fill=red!20] (Psi)
        {$\Psi_\mathcal{M}(\bar{\pi}) = c_{\bar{\pi}}^\ast$};
        % Draw edges
        \path [line,dashed] (A) -- (lproc);
        \path [line,dashed] (Pi) -- (lproc);
        \path [line,dashed] (C) -- (lproc);
        \draw[line] (lproc) to [out = -75, in = 65] (M);
        \draw[line,dashed] (M) to [out = -235, in = -105] (lproc);
        \path [line] (M) -- (recomm);
        \path [line,dashed] (pi) -- (recomm);
        \path [line] (recomm) -- (Psi);
        \path [line,dashed] (Psi) -- (lproc);
    \end{tikzpicture}
    \caption{Algorithmic schema of an ACP approach}
    \label{f:acp_schema}
\end{figure}

In all approaches in the literature, solving the ACP fits into the same two-stage framework (see Fig.~\ref{f:acp_schema}), encompassing the ordered execution of:
\begin{itemize}
 \item[a)] a {\it knowledge-encoding process} (for brevity, K-EP). The K-EP builds $\mathcal{M}$ and the accompanying $\Psi_\mathcal{M}$. A critical step in the computation of $\mathcal{M}$ is the sampling of the performance function, i.e., the evaluation of $p_\mathcal{A}$ over pairs $(\pi, c) \in \Pi' \times \mathcal{C}_\mathcal{A}$. Since $\mathcal{C}_\mathcal{A}$ may be quite large, in most cases computing $p_\mathcal{A}$ on all the configurations is too expensive. Thus, in all ACP approaches, the selection of an appropriate subset $\mathcal{C}'_\mathcal{A}$ of $\mathcal{C}_\mathcal{A}$ in the K-EP is a crucial task, which may require the use of $\mathcal{M}$ or additional models. Instead, when $\mathcal{C}_\mathcal{A}$ is small, all the $c \in \mathcal{C}_\mathcal{A}$ can be considered;
 \item[b)] a {\it recommendation phase}. The recommendation phase deploys $\Psi_\mathcal{M}$, in order to produce a suitable configuration for a given instance. Thus, $\Psi_\mathcal{M}$ supplies a solution of the ACP for that instance.
\end{itemize}

Since, as we noted above, most $\mathcal{M}$ are data-driven, the K-EP can demand considerable computational resources. The recommendation phase can also be computationally expensive, in that exploiting $\mathcal{M}$ to produce the output configuration may involve, e.g., the solution of a nontrivial optimization problem in itself.

The K-EP is an iterative procedure, run until an allotted computational budget
% $B$
(quantified, e.g., in terms of allowed target algorithm runs, CPU/GPU time and power, memory usage, number of K-EP iterations) is used up, or $\mathcal{M}$ has attained a desired accuracy. It cycles through three phases at each iteration $t \in \{0, \dots, T\}$:
\begin{enumerate}
\item {\sf sample$_t$}: picks a set
\begin{equation}
	\mathscr{S}_t\subset \Pi' \times \mathcal{C}_\mathcal{A} \quad~s.t.~\quad \mathscr{S}_t\ \cap\ \bigcup_{h < t} \mathscr{S}_h = \emptyset\,,
	\label{eq:acp_sample_set}
\end{equation}
i.e., $\mathscr{S}_t$ only contains previously unsampled couples. This selection is nontrivial, as it requires addressing the trade-off between uniformly exploring $\Pi' \times \mathcal{C}_\mathcal{A}$, so that no promising area is left unexplored (diversification), and concentrating on the areas containing the most promising candidates, so as to find better solutions (intensification);
\item {\sf evaluate$_t$}: executes (possibly, in parallel) the target algorithm $\mathcal{A}$ on all the points picked by {\sf sample$_t$}, to compute $p_\mathcal{A}$ at those points and build the set
\begin{equation}
\mathcal{S}_t = \big\{\, \big(\, \pi,\ c,\ p_\mathcal{A}(\pi, c)\, \big)\ |\ (\pi, c) \in \mathscr{S}_t\,\big\}\,;
\label{eq:acp_trset}
\end{equation}
\item {\sf update$_t$}: updates the models employed in the K-EP, i.e., the model $\mathcal{M}$ of $\Psi_\mathcal{M}$ and, potentially, other models used for sampling purposes. For instance, it may entail training or re-training an ML model. This phase exploits the set $\bigcup_{h \le t} \mathscr{S}_h$, the performance values computed at the points of that set and, sometimes, some other information collected in the ${\sf evaluate}_t$ phase. %Finally, {\sf update$_t$} checks whether any $B$ and $t$ are updated.
\end{enumerate}
The data generated by the recommendation phase may be employed in an adaptive ``meta-sampling'' loop whereby: a) when a new instance $\bar{\pi} \in \Pi \smallsetminus \Pi'$ is given (either by the user of the recommender, or by a dedicated process aimed at improving its quality), one computes $\Psi_\mathcal{M}(\bar{\pi})$; b) the recommended configuration and/or $\bar{\pi}$ are fed into the K-EP, which can be performed again to improve $\mathcal{M}$. One example of this approach is presented in \cite{clustering_asp_2}.

The implementation of the K-EP and the recommendation phase depends on the choice of the following components:
\begin{itemize}
 \item a model $\mathcal{M}$ and the associated recommender $\Psi_\mathcal{M}$;
 \item whether $\Psi_\mathcal{M}$ actually depends on a specific instance or always provides the same answer for a set of instances: we call {\it per-instance} ACP approaches of the former type, and {\it per-problem} approaches of the second type;
 \item whether one commits to building $\mathcal{M}$ before actually solving an unknown instance by $\mathcal{A}$ ({\it offline} ACP methodologies) or $\mathcal{M}$ is constructed during an algorithm run ({\it online} methodologies).
\end{itemize}

%---------------------------------------------
%---------------------------------------------
\section{Formulations}\label{s:acp_formulations}

%---------------------------------------------
\subsection{The construction of  $\mathcal{M}$}\label{ss:acp_m}

In the literature, the model $\mathcal{M}$ built in the K-EP is one of the following:
\begin{enumerate}[label=(\alph*)]
 \item[(a)] a function
\begin{equation}
\zeta_\mathcal{A}: \Pi \longrightarrow
                  \mathcal{C}_\mathcal{A}
\,,
\label{eq:m_ml_config}
\end{equation}
mapping an instance encoding to the configuration recommended for that instance. The function in Eq.~\eqref{eq:m_ml_config} is usually constructed by ML techniques \cite{optimal_trees_asp,learn_to_scale,zarp_miqp,paopai_lion};
 \item[(b)] a function
\begin{equation}
\bar{p}_\mathcal{A}: \Pi \times \mathcal{C}_\mathcal{A}
                \longrightarrow \mathbb{R}
\,,
\label{eq:m_ml_perf_fun}
\end{equation}
computing an approximation of the performance function $p_\mathcal{A}$ defined by Eq.~\eqref{eq:perf_fun}, also generally built by ML techniques \cite{piac,data_mining_acp,ehmhh,pmlp_cssp_lod}. Sometimes \cite{gga++}, $\bar{p}_\mathcal{A}$ is aggregated (e.g., averaged) over the instances in $\Pi'$, which yields an estimate
\begin{equation}
\chi_\mathcal{A}: \mathcal{C}_\mathcal{A}
             \longrightarrow \mathbb{R}
\label{eq:m_dist_pref_fun}
\end{equation}
of the performance of single configurations over $\Pi$. The functions $\bar{p}_\mathcal{A}$ or $\chi_\mathcal{A}$ are then used as a proxy to recommend a configuration for the new instance, typically, by solving an optimization problem having them in the objective;
\item[(c)] a partition
\begin{equation}
\mathscr{P}_\mathcal{A} = \{\Pi'_i \subseteq \Pi'\}_{i \in C}
\label{eq:m_partition}
\end{equation}
of $\Pi'$ into $C$ disjoint subsets (or ``clusters'') $\Pi'_i$, whereby each $\Pi'_i$ is specified by choosing the corresponding recommended configuration ${\bar{c}^*_i}$ and, for instance, a representative instance $\pi_i$. When a new instance $\bar{\pi} \in \Pi$ has to be solved, the cluster to which it belongs is determined (e.g., by finding the closest representative $\pi_i$, under some appropriate distance metric) and the corresponding $c^*_i$ is retrieved.

We remark that there are two ``extreme'' cases of Eq.~\eqref{eq:m_partition}:
\begin{itemize}
\item one in which $C = |\Pi'|$, i.e., each $\Pi'_i$ contains exactly one instance, \cite{reactive,bayesian_cbr}. We refer to this case as ``$\mathscr{P}_{\mathcal{A}, |\Pi'|}$'';
\item one whereby $\Pi'_0 = \Pi'$ and $C = 1$, i.e., the partition is trivial and a single configuration ${\bar{c}^*_0}$ will be recommended regardless of the input $\bar{\pi}$. This is customary in per-problem approaches \cite{calibra,gga,opal_2,opal_1,paramils_vns,gga++,smac,paramils,revac,irace}. We refer to this case as ``$\mathscr{P}_{\mathcal{A}, 1}$''.
\end{itemize}
Moreover, we refer to intermediate case, whereby $1 < C < |\Pi'|$,  as ``$\mathscr{P}_{\mathcal{A}, C}$''.

It should be noted that this choice of $\mathcal{M}$ is typically strongly coupled with the sample$_t$ phase in the K-EP, in the sense that it is often the direct result of the algorithmic decisions there.

The strategy implemented to construct $\mathscr{P}_\mathcal{A}$ is that of solving the problem
\begin{equation}
{\bar{c}^\ast_i}	 =
\textstyle
\argmax \{ \, \agg_{\pi \in \Pi_i}p_\mathcal{A}(\pi, c)
         \,|\, c \in \mathcal{C}_\mathcal{A} \,\}\,,
\label{eq:min-per-cluster}
\end{equation}
where $\agg$ is some aggregation function (say, the average) and we want to maximize $p_\mathcal{A}$. The issue here is to find the configuration providing the best aggregated performance with respect to the subset $\Pi'_i$ at hand. Since $\mathcal{C}_\mathcal{A}$ is often very large and $p_\mathcal{A}$ is ``black-box'', the problem in Eq.~\eqref{eq:min-per-cluster} is typically treated by heuristic search algorithms, such as local searches, evolutionary algorithms or other metaheuristics, providing a local solution. Further, such heuristic algorithms can themselves naturally identify clusters on the fly, even in the extreme cases; for instance, evolutionary algorithms produce a population of the fittest individuals, each of which can be used to define a cluster representative $\pi_i$.

We remark that the strategy of building $\mathscr{P}_\mathcal{A}$ by solving Eq.~\eqref{eq:min-per-cluster}, usually adopted by offline ACP methodologies, quite naturally extends to the online setting. In the online ACP, further evaluations of $p_\mathcal{A}$ are often triggered by the arrival of a new $\pi$, that restarts the optimization process.
\end{enumerate}
Some ACP methodologies combine more than one model; the possible combinations are shown in Tab.~\ref{acp_overview}.

%---------------------------------------------
\subsection{Computing $\Psi_\mathcal{M}$}\label{ss:acp_psim}

Given an instance $\bar{\pi} \in \Pi$, the implementation of $\Psi_\mathcal{M}(\bar{\pi})$ depends on the model $\mathcal{M}$ built during the K-EP:
\begin{enumerate}[label=(\alph*)]
\item If $\mathcal{M}$ is the one in Eq.~\eqref{eq:m_ml_config}, then
	   $\Psi_\mathcal{M}(\bar{\pi}) := \zeta(\bar{\pi})$;
 \item if $\mathcal{M}$ is the one in Eq.~\eqref{eq:m_ml_perf_fun} and we want to maximize algorithmic performances, then
\begin{equation}
 \textstyle
 \Psi_\mathcal{M}(\bar{\pi}) :=
 \arg\max \{ \, \bar{p}_\mathcal{A}(\bar{\pi}, c)
                  \,|\, c \in {\mathcal{C}'_\mathcal{A}} \,\}
    \,.
	\label{eq:cssp_barp_instance_single}
\end{equation}
The problem in Eq.~\eqref{eq:cssp_barp_instance_single} may be a hard one, calling for heuristic solution algorithms akin to those possibly employed in the K-EP, i.e., treating the approximation $\bar{p}_\mathcal{A}$ as a ``black box'' (e.g., \cite{piac,ehmhh,hydra_1,hydra_2,gga++,autosklearn_1}). However, exploiting the mathematical structure of $\bar{p}_\mathcal{A}$ is also possible, allowing the use of exact approaches \cite{pmlp_cssp_lod};
\item if $\mathcal{M}$ is specified by a set of clusters $\Pi'_i \subseteq \Pi'$ as in Eq.~\eqref{eq:m_partition}, then $\Psi_\mathcal{M}(\bar{\pi})$ is implemented by first solving
 \begin{equation}
\textstyle
	h(\bar{\pi}) = \mbox{argmin}_i \, {\sf dist}(
	%{\cancel{\Pi'_i}}
	{\pi_i} , \bar{\pi} )
 \,,
	\label{eq:findrep}
\end{equation}
for some appropriate distance function ${\sf dist}(\cdot,\cdot)$ {and a representative instance $\pi_i$ specifying the cluster $\Pi'_i$}, and then returning $c^*_{h(\bar{\pi})}$. {A more expensive alternative would be to compute the average $\mathsf{dist}$ between the instances in $\Pi_i$ and $\bar{\pi}$.
%\sout{If the clusters are specified via a representative instance, say, $\pi_i$ for cluster $\Pi_i$, then}
}
%\[{\cancel{{\sf dist}( \Pi'_i , \bar{\pi} ) = \| \bar{\pi} - \pi_i \|\,,}}\]
%{\sout{for some appropriate norm $\|\cdot\|$}}.
When the number of clusters is low, as it usually happens, Eq.~\eqref{eq:findrep} can be quickly solved by direct enumeration. In the extreme case with only one cluster, instead, the problem is trivial.
\end{enumerate}

%---------------------------------------------
\subsection{Classifying algorithm configuration approaches}\label{s:acp_pp_pi_off_on}

Per-problem approaches search for the configuration with the best overall performance over a problem set; they are usually based on one of the models $\mathcal{M}$ described at point (c) of Sec.~\ref{ss:acp_m}. The main risk of per-problem approaches is that they produce suboptimal ACP solutions when the performance of the target algorithm varies considerably between instances of a problem. This is to be expected for large problem classes, e.g.~mixed-integer linear programming, where instances can be hard for very different reasons (say, having very few feasible solutions, so that finding even one is challenging, or very many feasible solutions with very close objective, so that finding the optimal one is challenging). In this case, per-instance methodologies, which assume that the ACP-optimal algorithmic configuration depends on the instance at hand, are likely to achieve better results.

An ACP methodology is offline if it commits to building $\mathcal{M}$ before the recommendation phase, during which $\mathcal{M}$ is, thus, fixed. Instead, an online methodology performs the entire K-EP during the execution of $\mathcal{A}$. In this case, recommendation phase and K-EP coincide, and the information retrieved by running $\mathcal{A}$ is dynamically exploited, on-the-fly, to build $\mathcal{M}$.
Online methods can only be employed when the target algorithm is launched to solve a sequence of instances, or is tasked with making a series of decisions during its run.

An online procedure can be used as a component of a larger offline/online approach. One way to accomplish this would be to construct, online, $\mathcal{M}$ through experiments on $\Pi'$, and, then, deploy it as a recommender for instances similar to those in $\Pi'$.
A very straightforward implementation of this approach would be to, say, run an optimization solver on a sequence of instances $\Pi'$, within a prescribed time limit, trying different parameter configurations on each of them. This would allow the selection and storage of a set of parameter values ensuring high solver performance (as in model $\mathscr{P}_\mathcal{A}$ in Eq.~\eqref{eq:m_partition}); they could be reused to configure the solver and run it on different instances in $\Pi$. Several commercial optimization solvers (e.g., Gurobi \cite{gurobi} or CPLEX \cite{cplex127}) have automatic tuning tools that work in this way and can be used to implement this procedure.

Another way to reuse an online $\mathcal{M}$ could be to employ it in the {\sf update$_0$} phase of a subsequent K-EP, to initialize the construction of a new model. Yet another option may be to use the points (instance, configuration, related performance), sampled to construct $\mathcal{M}$ online (by the {\sf sample}$_t$ and the {\sf evaluate}$_t$ phases), in the {\sf sample$_0$} phase of a following K-EP.

Since the purpose of online approaches is to solve the ACP on-the-run, they are usually based on simple algorithms, which enable rapid decision-making. However, these approaches are often impractical to scale to large configuration sets. For this reason, they are ordinarily used for solving the ASP, rather than the ACP.

\begin{table}[h]
\centering
\scalebox{1.2}{
\begin{tabular}{c|c|c|c}
& \multicolumn{2}{c|}{\hyperlink{acronyms_anchor}{per-instance}} & \multicolumn{1}{c}{\hyperlink{acronyms_anchor}{per-problem}} \\
\hline
$\mathcal{M}$ & offline & online & offline \\
\hline
$\zeta_\mathcal{A}$ (Eq.~\eqref{eq:m_ml_config})
& \cite{optimal_trees_asp}$^\ast$,
\cite{learn_to_scale,zarp_miqp,paopai_lion} & & \\
\hline
$\bar{p}_\mathcal{A}$ (Eq.~\eqref{eq:m_ml_perf_fun}) & \cite{piac,data_mining_acp,ehmhh,pmlp_cssp_lod} & & \\
\hline
$\mathscr{P}_{\mathcal{A}, C}$ (Eq.~\eqref{eq:m_partition}) &
\cite{clustering_asp_1,clustering_asp_2}$^\ast$, \cite{isac} & & \\
\hline
$\mathscr{P}_{\mathcal{A}, 1}$ (Eq.~\eqref{eq:m_partition}) & & \cite{reactive} &
\shortstack{\cite{calibra,gga,opal_2,opal_1,paramils_vns}\\\cite{paramils,revac,irace}} \\
\hline
$\mathscr{P}_{\mathcal{A}, |\Pi'|}$ + $\chi_\mathcal{A}$ (Eq.~\eqref{eq:m_partition} + \eqref{eq:m_dist_pref_fun}) & & \cite{bayesian_cbr} & \\
\hline
$\mathscr{P}_{\mathcal{A}, 1}$ + $\bar{p}_\mathcal{A}$ (Eq.~\eqref{eq:m_partition} + \eqref{eq:m_ml_perf_fun})&
\cite{autosklearn_1,hydra_1,hydra_2} & \cite{autosklearn_1} & \cite{smac}\\
\hline
$\mathscr{P}_{\mathcal{A}, 1}$ + $ \chi_\mathcal{A}$ (Eq.~\eqref{eq:m_partition} + \eqref{eq:m_dist_pref_fun})&
& & \cite{gga++}
\end{tabular}
}
\caption{A schematic summary of the ACP literature; $^\ast$ indicates ASP approaches.
\label{acp_overview}}
\end{table}

In Tab.~\ref{acp_overview}, we give an overview of the main ACP approaches in the literature.
From the rightmost column of the table, we gather that all per-problem methodologies are based on the model $\mathscr{P}_\mathcal{A}$ described by Eq.~\eqref{eq:m_partition}, notably, on its $\mathscr{P}_{\mathcal{A}, 1}$ variant. Instead, the $\mathscr{P}_{\mathcal{A}, C}$ and $\mathscr{P}_{\mathcal{A}, |\Pi'|}$ variants are always used in the per-instance setting.
Further, most per-instance approaches rely on the ML-derived models $\bar{p}_\mathcal{A}$ of Eq.~\eqref{eq:m_ml_perf_fun} and $\zeta_\mathcal{A}$ of Eq.~\eqref{eq:m_ml_config}.
In fact, per-instance methodologies are required to capture/encode the complicated, possibly nonlinear relationships between $p_\mathcal{A}$, $\mathcal{C}_\mathcal{A}$ and $\Pi$, and several ML paradigms are capable of producing extremely accurate approximation.
In some cases, ML-based models and variants of $\mathscr{P}_\mathcal{A}$ are combined, in order to implement per-instance procedures: see, e.g., the approaches on the second to last line of the table, which rely on both $\mathscr{P}_{\mathcal{A}, 1}$ and $\bar{p}_\mathcal{A}$ for online and offline algorithm configuration.

%---------------------------------------------
%---------------------------------------------
\section{Conclusions}\label{s:acp_conclusions}

We introduced an algorithmic schema, common to all ACP methodologies, for constructing and deploying a recommender, i.e., a function capable of suggesting the optimal configuration of a given algorithm $\mathcal{A}$ for solving an instance of a given decision/optimization problem $\Pi$. %Furthermore, we have defined a taxonomy of the ACP.

Despite all the research efforts in the field of algorithm configuration, the problem still remains extremely difficult to solve. In fact, it is usually impossible to know the algorithmic performance $p_\mathcal{A}$ over all of the many parameter configurations in $\mathcal{C}_\mathcal{A}$, which can be { in the order of thousands or much larger} (especially in general-purpose optimization solvers, equipped with a diverse set of algorithmic components to solve famously NP-hard problems \cite{npcompl_milp,nlp_nphard}).
Furthermore, while the ACP can be somewhat {
%\sout{approached}
manageable} if we consider a single instance or a small subset of $\Pi'$, it becomes intractable when we look at the whole set $\Pi'$, which is normally of infinite size.

To overcome this complication, ACP methodologies are all based on multiple forms of approximation of: a) the algorithmic performance function, or of the parameter configuration allowing to achieve specific algorithmic performances, via some computable ML approximation; b) $\Pi$, via the manual selection of a subset of representative instances and corresponding representative configurations.

% ---- Bibliography ----
\ifarxiv
\bibliographystyle{unsrt}%apalike
\else
\bibliographystyle{splncs04}
\fi
% use tiny in case of need
{\bibliography{../biblio_giommazz}}

\clearpage
\appendix
\onecolumn
\end{document}